\documentclass[letterpaper, 10pt, conference]{IEEEtran}
\usepackage{times}
\usepackage{xr}
\usepackage{amsmath,amssymb,mathrsfs,gensymb}
\usepackage[]{algorithm2e}
\usepackage{graphicx}
\usepackage[dvipsnames]{xcolor}
\usepackage{wrapfig}
\usepackage{textcomp}
\usepackage{duckuments}

\usepackage{booktabs}

\definecolor{nicegreen}{RGB}{80, 200, 120}

\definecolor{custompink}{RGB}{255, 154, 154}

\usepackage[sort,numbers,compress]{natbib}
\usepackage{multicol}
\usepackage[bookmarks=true, hidelinks]{hyperref}

\begin{document}

\title{
\LARGE\bf
Creating manufacturable blueprints for
coarse-grained virtual robots
}

\author{Zihan Guo$^*$, Muhan Li$^*$, Shuzhe Zhang, Sam Kriegman\\
Northwestern University}




\maketitle

\begin{abstract}
Over the past three decades, countless embodied yet virtual agents have freely evolved inside computer simulations,
but vanishingly few were realized as physical robots.
This is because evolution was conducted at a
level of abstraction that was convenient for
freeform 
body generation (creation, mutation, recombination) but
swept away almost all of the physical details of functional body parts.
The resulting designs were crude and underdetermined,
requiring considerable effort and expertise to convert 
into a manufacturable format.
Here, we automate this mapping from simplified design spaces that are readily evolvable to complete blueprints that can be directly followed by a builder.
The pipeline
incrementally resolves manufacturing constraints  
by embedding the structural and functional semantics of motors, electronics, batteries, and wiring into the abstract virtual design.
In lieu of evolution, a user-defined or AI-generated ``sketch'' of a body plan can also be fed as input to the pipeline,
providing a versatile framework for accelerating the design of novel robots.
\end{abstract}

\IEEEpeerreviewmaketitle

\begin{figure*}[!t]  
  \centering
  \includegraphics[width=\textwidth]{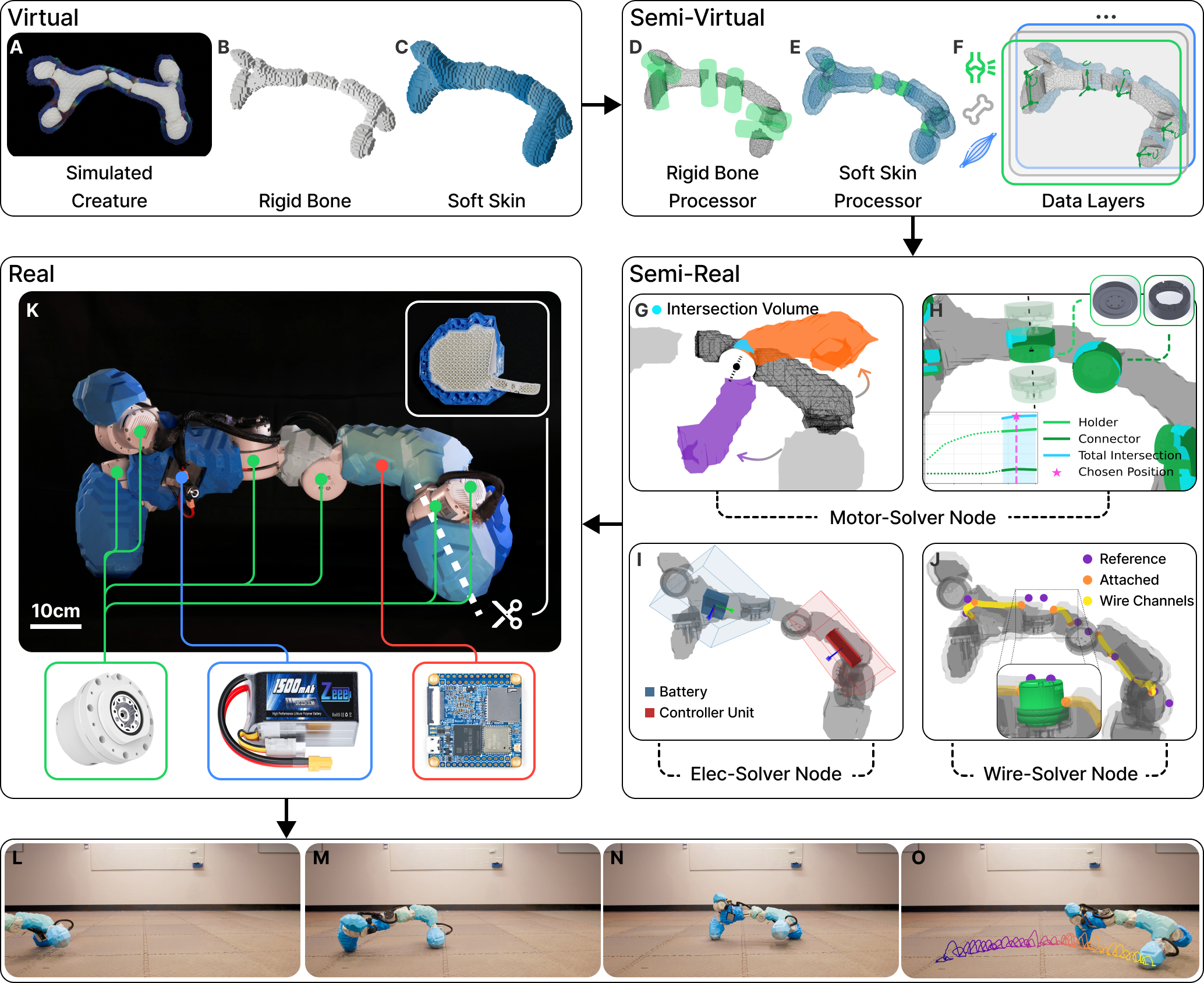} 
  \caption{\textbf{Converting high-level abstractions into real robots.} 
  The pipeline begins in simulation (virtual stage; \textbf{A-C}) where abstract, voxelized robots---``virtual creatures''---can freely evolve complex endoskeletal morphology without recourse to real world manufacturing constraints.
  A design with high fitness in simulation is shown in A.
  This voxelized representation is fed as input to the ``semi-virtual'' stage for processing (\textbf{D-F}).
  Here, the rigid bone processor converts rigid voxels to bone meshes (gray surfaces in D) 
  and cuts cylinders (green rolls in D) along the joint axis to reserve space for motors. 
  The soft skin processor cleans space around joints (green spheres in E), 
  applies erode-dilate smoothing 
  and converts soft voxels into hollow skin meshes (blue surfaces in E) to avoid self-collision. 
  The output of these processors is composed into a layered data structure (F) and forward propagated to the next stage.
  %
  We refer to the next part of the pipeline
  the ``semi-real'' stage because it
  uses solver nodes to
  incrementally resolve manufacturability constraints,
  but the interim design is neither realized nor a fully fledged blueprint.
  First, admissible joint motion ranges are estimated by the motor solver (\textbf{G}) and candidate motor holder/connector depths are scanned along each joint axis to select the optimal motor position (\textbf{H}). 
  Next, the electronics solver carves cavities for the controller and battery installation box
  into the two body parts with the largest volume (\textbf{I}). 
  Wire tunnels are then routed (\textbf{J}) through both rigid and soft meshes by the wire solver.
  The resulting assembly-ready multi-material body parts are then 3D printed
  and assembled in the real world (\textbf{K}).
  A cross-section of one of the endoskeletal body parts is displayed (in K).
  The evolved tripedal design successfully transferred to reality, exhibiting forward locomotion under the learned closed-loop policy (\textbf{L-O}).
  The body rocks back and forth, pivoting on single hind limb as the left and right forelimbs alternate contact with the ground.
  The trajectory of the right forelimb is traced (from purple to yellow) in O.
  }
  \vspace{-1em}
  \label{fig:pipeline-overview}
\end{figure*} 
\begin{figure*}[!t]  
  \centering
  \includegraphics[width=\textwidth]{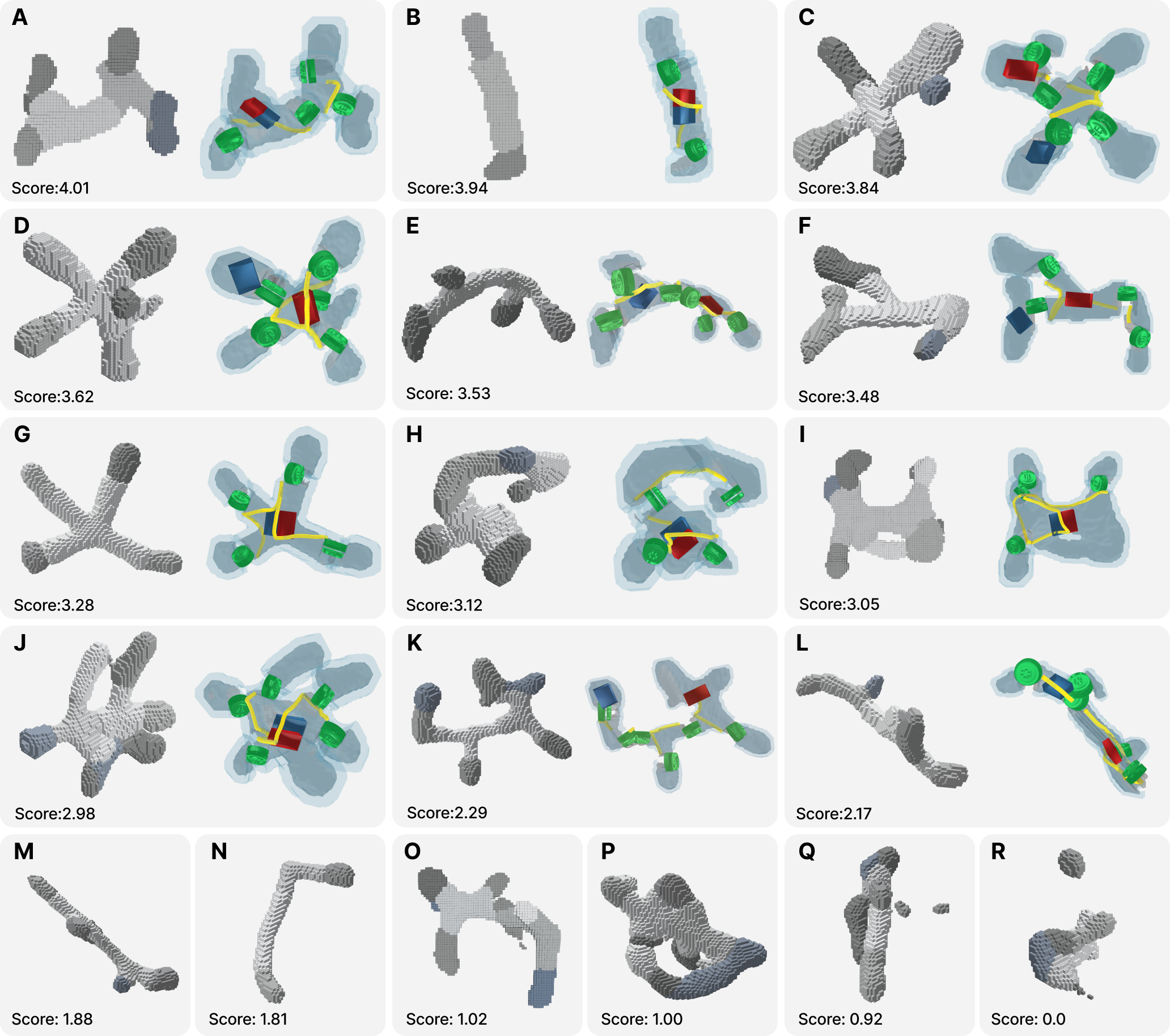} 
  \caption{\textbf{Running the pipeline.} 
  We tested our pipeline against a large, diverse set of designs randomly sampled from the latent space,
  a representative subset of which are displayed (\textbf{A-R}).
  For each design, 
  motors (green cylinders), a battery (blue boxes), and a controller box (red boxes) are automatically positioned and connected by wire tunnels (yellow lines).
    The resulting anatomy is then assigned a manufacturability score based on
    the estimated feasibility/strength of the selected motor positions, 
    the availability of free space for the electronics and battery, 
    the curvature necessary for cable tunnels,  
    the clearance above the electronics box,
    and collision of body parts after motor integration.
    Low-quality designs have lower scores.
    For example, designs 
    that 
    have very long cable tunnels (I),
    are too thin for the electronics box (M, N),
    lack the spacing between joints for the motors (K, L),
    or 
    possess an invalid kinematic tree (Q, R).
  }
  \vspace{-1em}
  \label{fig:multiple_morphologies}
\end{figure*}

\begin{figure}[!t]
  \centering
  \includegraphics[width=\columnwidth]{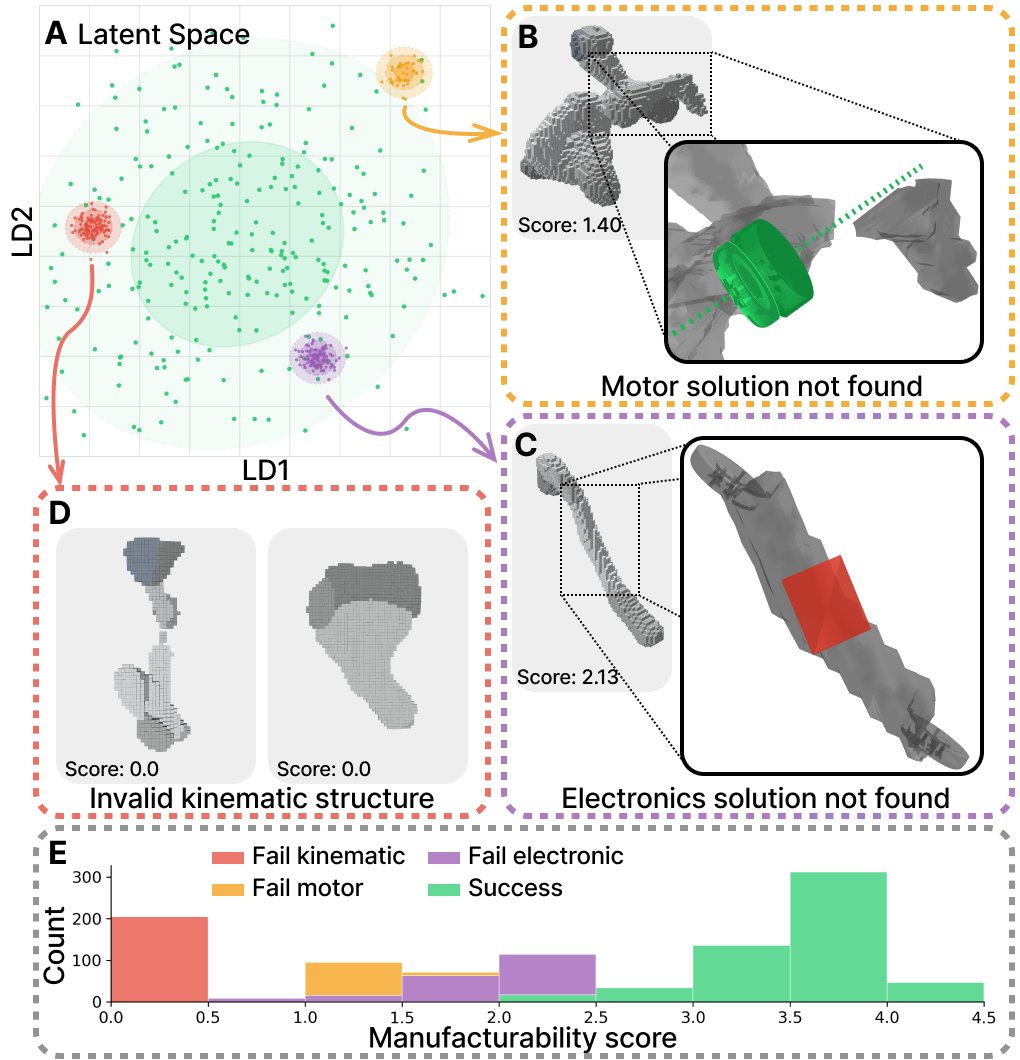}
  \caption{\textbf{Failure modes.}
  Designs were sampled randomly from the latent space, which can be projected by linear discriminant analysis  into a 2D map (\textbf{A}).
  Each axis is a linear combination of latent features,
  and shows that failures from the motor (\textbf{B}) and electronics (\textbf{C}) solvers and the rigid bone processor (\textbf{D}) 
  can be isolated and avoided during design evolution. 
  A histogram of the manufacturability scores from sampled designs reports the number of robots within each score interval (\textbf{E}). 
  } 
  
  \vspace{-1em}
  \label{fig:failure}
\end{figure}

\section{Introduction}
\label{sec:intro}

Inspired by the early 1990s simulations of \citet{sims1994competition}
and
\citet{ventrella1994explorations}, 
\citet{lipson2000automatic} evolved virtual creatures in a simulated environment and then manufactured the most promising evolved designs as physical robots.
Since then many others \cite{hiller2012automatic,cellucci20171d,kriegman2020xenobots,kriegman2020scalable,kriegman2021kinematic,kriegman2021fractals,moreno2021emerge,schaff2022soft,matthews2023efficient,strgar2024evolution,kobayashi2024computational,yu2025reconfigurable} have followed in their footsteps, 
automating the search for good body plans and behaviors in simulation before building and testing transferable designs in reality \cite{kriegman2019virtual}.

A variety of methods have been developed for modeling the behavior of evolving body plans, but they typically rely on highly abstract virtual representations (e.g.~particles \cite{matthews2023efficient,wang2023diffusebot},
springs \cite{strgar2024evolution,strgar2025accelerated}, or  voxels~\cite{hiller2012automatic,cheney2018scalable}) 
that lack any manufacturable semantics of actuators, electronics, batteries, or wires.
Removing these physical details simplifies the generation, mutation, and recombination of freeform morphological structures,
and reduces the amount of time required to evaluate the fitness of each candidate design.
But, in order to realize any one of these abstract designs as a physical robot,
a human expert was needed to manually reinterpret the simulated body, create a detailed CAD model, build physical prototypes and iterate several times.
This process is labor-intensive, time-consuming, and fundamentally limits the scalability of evolutionary robotics.

While several efforts have sought to
reduce the manual burden of transferring evolved designs from simulation to reality, these approaches have operated in highly constrained design spaces, such as those that 
predefine the robot's kinematic structure~\cite{ringel2025text2robot,xu2021end},
restrict the body plan to 
a small collection of simple primitives (e.g.~rigid cylinders~\cite{lipson2000automatic}),
constrain designs to one~\cite{cellucci20171d} or two~\cite{strgar2024evolution,matthews2023efficient} dimensional spaces,
or utilize large predesigned modules~\cite{schaff2022soft, moreno2021emerge, brodbeck2015morphological, yu2025reconfigurable}.
While these strategies can reduce the virtual-to-real gap, they also restrict the design space and thus limit the diversity and functional complexity of the possible design innovations that could evolve.

Here we introduce a scalable framework for automatically translating abstract virtual agents with freeform morphology into assembly-ready, manufacturable designs. 
We show results for complex endoskeletal body plans that combine internal articulated skeletons with external soft tissues~\cite{li2025generating}. 
Central to our approach is viewing the virtual-to-real process as a task decomposition problem: the pipeline incrementally resolves manufacturing constraints by embedding manufacture relevant semantics 
(e.g.~those of motors, electronics, batteries, and wiring) into the abstract representation. 
This eliminates the need for human intervention during CAD redesign, 
enabling coarse-grained designs
to be realized directly in real hardware. 

\section{Methods}
\label{sec:methods}


Our goal here is to convert
virtual creatures 
with arbitrary structural and architectural features
into
physical functional robots using a given set of parts and materials.
We introduce a pipeline (Fig.~\ref{fig:pipeline-overview}) that decomposes this problem into four stages, which we will refer to as virtual (Sect.~\ref{subsec:virtual}), 
semi-virtual (Sect.~\ref{subsec:semi-virtual}), 
semi-real (Sect.~\ref{subsec:semi-real}), 
and real (Sect.~\ref{subsec:real}).
As the design progresses through these stages
it incrementally gains the
semantics and components
necessary to output a complete blueprint that can be directly followed by a builder.

\subsection{Virtual}
\label{subsec:virtual}

The pipeline begins with a
freeform virtual creature that exhibits promising behavior in simulation (e.g.~the design in Fig.~\ref{fig:pipeline-overview}A).
Here, we adopt the 
endoskeletal design space and
latent genetic compression from \citet{li2025generating}. 
Thus designs are abstract yet relatively complex: 
they contain 
freeform 
skeletons with arbitrary bone shapes
(Fig.~\ref{fig:pipeline-overview}B) 
and soft tissues (Fig.~\ref{fig:pipeline-overview}C), but they consist purely of voxels and joints.
More specifically, a latent genotype vector $\mathbf{z}\in\mathbb{R}^{512}$ is sampled from the latent space 
and decoded onto a voxel grid $\mathcal{G}\in\mathcal{L}^{64\times 64\times64}$
as rigid and elastic voxels.
The virtual representation also includes a set of joint estimates
$\mathcal{J}=\{(\mathbf{p}_j, \mathbf{a}_j, \Phi_j)\}$ 
that specifies each joint's position $\mathbf{p}_j$, axis $\mathbf{a}_j$, and range of motion $\Phi_j$. 
This voxelized morphology and its kinematic annotations are fed as input to the next stage of the pipeline for processing.

\subsection{Semi-virtual}
\label{subsec:semi-virtual}

We refer to this stage as
``semi-virtual'' 
because it
produces fabrication-oriented geometric representations  (Fig.~\ref{fig:pipeline-overview}D-F) for the virtual morphology but does not directly include specific hardware constraints. 
Voxels are intuitive and convenient for generative modeling and simulation, but they are not directly compatible with the CAD and manufacturing ecosystem, 
and they provide limited support for precise geometric operations. 
The pipeline uses two processors, one for each material type (soft and rigid), to convert the voxel representation into triangle meshes, which serve as the common geometry interface for downstream solvers and enable 
exporting 
as standard fabrication files.

Mesh geometry also provides a natural substrate for representing hierarchical design detail. 
Starting from the global morphology surface, the same mesh representation can be progressively augmented with localized semantics---including actuator and electronics accommodation sites, 
internal wire-routing paths, 
and fine fabrication features as from screw holes and chamfers, using standard mesh processing operators such as boolean union/difference.

Given the decoded voxel grid $\mathcal{G}$, two complementary meshes are generated: rigid bone meshes and soft skin meshes. 
For each rigid body segment, 
a rigid-bone processor converts the segment occupancy into a triangle mesh representation using Marching Cubes~\cite{Lorensen1987MarchingCA}. 
To make joint regions explicit and easier to manipulate in downstream stages, the bone processor subtracts a cylindrical clearance volume centered at each joint position $\mathbf{p}_j$ and oriented along the joint axis $\mathbf{a}_j$ (Fig.~\ref{fig:pipeline-overview}D). 
For soft voxels, a skin processor first removes voxels within a fixed radius of each joint to create a local clearance region, then an erode-dilate operation is applied to smooth sharp geometric features around the joint neighborhoods (Fig.~\ref{fig:pipeline-overview}E) before extracting a continuous skin mesh via Marching Cubes. 
Finally, the processed meshes and the kinematic annotations are compiled into a unified data format (Fig.~\ref{fig:pipeline-overview}F).

\subsection{Semi-real}
\label{subsec:semi-real}

We call this part of the pipeline ``semi-real'' because it converts the semi-virtual geometry into an assembly-ready design by explicitly integrating real hardware components (Fig.~\ref{fig:pipeline-overview}G-J), but the design remains digital and the blueprint incomplete until this stage terminates. 
We conceptualize this stage as a constraint-solving problem with structured interdependencies between different constraints. 
We therefore factor this stage into a sequence of solver nodes, each of which resolves a unique manufacturing constraint while updating the current design state. 
Each solver node $\mathcal{N}$ has the form:
\begin{equation}
\mathcal{O} = \mathcal{N}(\mathcal{I}, \theta),
\end{equation}
where $\mathcal{I}$ denotes the node input 
(e.g.~mesh and joint data), 
$\mathcal{O}$ denotes the node output 
(e.g.~updated meshes with embedded features or component position) 
and $\theta$ specifies the set of hyperparameters that define the manufacturing constraint for current node. 
If a node cannot find a feasible solution, it returns a failure reason and the pipeline terminates; 
otherwise, its output becomes the input to the next node.

\textbf{Motor solver node.}
The motor solver converts each abstract joint into a motor mounting and coupling structure while enforcing two key constraints. 
The input $\mathcal{I}_{\text{motor}}$ includes the cleaned rigid-body meshes and joint annotations $\mathcal{J}=\{(\mathbf{p}_j,\mathbf{a}_j,\Phi_j)\}$ from the semi-virtual stage, and the predesigned motor holder/connector mesh 
(see Fig.~\ref{fig:pipeline-overview}H).
The output $\mathcal{O}_{\text{motor}}$ is an updated set of rigid meshes with embedded motor holder/connector geometry along with the motor placement transforms for assembly. 
There are two key constraints for this node.
First, because simulated rigid segments might interpenetrate and ignore self-collisions (Fig.~\ref{fig:pipeline-overview}A),
feasible joint motion ranges are computed
and
structural interference near joint limits (i.e.~the mutual intersection of the jointed pair) is carved out
yielding a collision-reduced geometry that is easier to assemble and outfit with motors (Fig.~\ref{fig:pipeline-overview}G).
Second, the motor is placed at an estimated optimal offset along the joint axis $\mathbf{a}_j$ by scanning candidate positions and selecting the offset that provides a strong and balanced attachment to both connected bodies (Fig.~\ref{fig:pipeline-overview}H). More precisely, for each joint connecting rigid body segments $(s_1,s_2)$, we evaluate two assignment configurations $c$ (holder on $s_1$ and connector on $s_2$, or vice versa) and choose the offset $\delta^\star$ and configuration $c^\star$ that maximize the volume-based score:
\begin{equation}
S(\delta)=\mathbb{I}\!\left[V_h(\delta)\ge\tau,\,V_c(\delta)\ge\tau\right]\; g(\delta),
\label{eq:motor_score}
\end{equation}
with 
\begin{equation}
g(\delta)=\sqrt{V_h(\delta)V_c(\delta)}+\alpha\big(V_h(\delta)+V_c(\delta)\big),
\end{equation}
where
$V_h(\delta)$ and $V_c(\delta)$ denote the intersection volume between the holder/connector (placed at $\mathbf{p}_j+\delta \mathbf{a}_j$) and their target bodies. 
In Fig.~\ref{fig:pipeline-overview}H, the light and dark green curves plot $V_h(\delta)$ and $V_c(\delta)$, respectively, as $\delta$ is scanned along the joint axis,
and the light blue curve in Fig.~\ref{fig:pipeline-overview}H shows $V_h(\delta)+V_c(\delta)$.
The minimum-volume threshold
$\tau$ is used to reject weak attachments.
The weight $\alpha$ controls the trade-off between balanced attachment to both bodies and overall contact volume.
The selected offset $\delta^\star$ (pink star in Fig.~\ref{fig:pipeline-overview}H) maximizes the score $S(\delta)$, balancing strong attachment to both connected segments. 
Once $(\delta^\star,c^\star)$ is selected,
the design is finalized with mesh boolean union/difference operations, carving the required clearances and screw-hole patterns and integrating the holder/connector geometry.

\textbf{Electronics solver node.}
The electronic solver embeds accommodations for the battery and controller units into rigid body parts that are large enough to accept them (Fig.~\ref{fig:pipeline-overview}I). 
The input $\mathcal{I}_{\text{elec}}$ consists of the updated rigid meshes after the motor solver (with the motor mounts), together with the predefined geometries for the battery and controller box. 
The output $\mathcal{O}_{\text{elec}}$ is an updated set of rigid meshes with carved out cavities plus the global position and relative orientation of the controller unit.
The core constraint is containment with clearance: the electronics must lie inside the target rigid body and must not collide with previously embedded motor geometry. 
The solver uses a simple and robust placement strategy: for each candidate rigid segment $s$, the position is anchored at its center of mass, $\mathbf{p}_{\text{elec}}=\mathrm{CoM}(s)$, while scanning a discrete set of orientations $\mathbf{R}_{\text{elec}}\in SO(3)$; a candidate placement $(\mathbf{p}_{\text{elec}}, \mathbf{R}_{\text{elec}})$ is accepted if the geometries box is sufficiently contained within the segment and remains collision-free with the existing internal features.
If no single segment can host both the battery and controller, they are distributed across the largest available segments under the same containment test.
Once a feasible placement is found, the design is committed by carving out the cavities and recording the final electronics component transforms for downstream wiring and assembly.

\textbf{Wire solver node.}
The wire solver generates internal wire channels that connect each motor to the controller (Fig.~\ref{fig:pipeline-overview}J). 
The input $\mathcal{I}_{\text{wire}}$ includes the rigid meshes after the motor- and electronic solvers, together with the set of motor poses $(\mathbf{p}_j+\delta^\star \mathbf{a}_j)$ and electronics poses $(\mathbf{p}_\text{elec}, \mathbf{R}_\text{elec})$ that define the required connectivity endpoints. 
The output $\mathcal{O}_{\text{wire}}$ updates the rigid and soft meshes with embedded wire tunnels.
In order to determine the start and end points for cable path, a set of reference routing points are generated (purple dots in Fig.~\ref{fig:pipeline-overview}J) by offsetting from each motor and electronics frame, which are then snapped to the nearest mesh vertices to obtain the actual connecting points along mesh surface (orange dots). 
Geodesic paths are then computed along the mesh surface between the corresponding start-end pairs and 
smoothed by a moving-average filter over the polyline vertices.
Wire tunnels are created by sweeping a radius $r_\text{wire}$ tube along the smoothed paths (yellow lines in Fig.~\ref{fig:pipeline-overview}J). 
After selecting a feasible set of paths, the tunnels are subtracted from the rigid/soft meshes.

\subsection{Real}
\label{subsec:real}

At the end of the semi-real stage, the pipeline outputs a 
manufacturable blueprint: 
all rigid bodies contain motor mounts, electronics cavities, and internal cable tunnels, and all soft components are shaped to accommodate joint clearances and embedded hardware. 
These meshes can be directly exported as standard fabrication files (STL) and sent to the 3D printer without additional human CAD modification. 
Now we fabricate and assemble the robot according to the generated design.

\textbf{Mechanical fabrication.}
All structural components were fabricated using dual-material fused-filament fabrication. 
Rigid parts were printed using PLA, 
and compliant skin elements were printed using TPU (Shore-A 90), 
enabling integrated rigid-soft structures within a single build. 
We used a Bambu H2D dual-material printer with a 0.6\,mm nozzle. 
To improve interfacial bonding between PLA and TPU, we introduce a material interlocking depth of 0.63\,mm across rigid-soft boundaries. 
Both materials use a gyroid infill pattern to achieve near-isotropic in-plane stiffness while maintaining structural strength. 
To approximate the compliance assumed in simulation, soft skin components were printed with 5\% infill and a single perimeter wall, minimizing shell-induced stiffening while preserving printability.

\begin{figure}[!t] 
  \centering
  \includegraphics[width=\columnwidth]{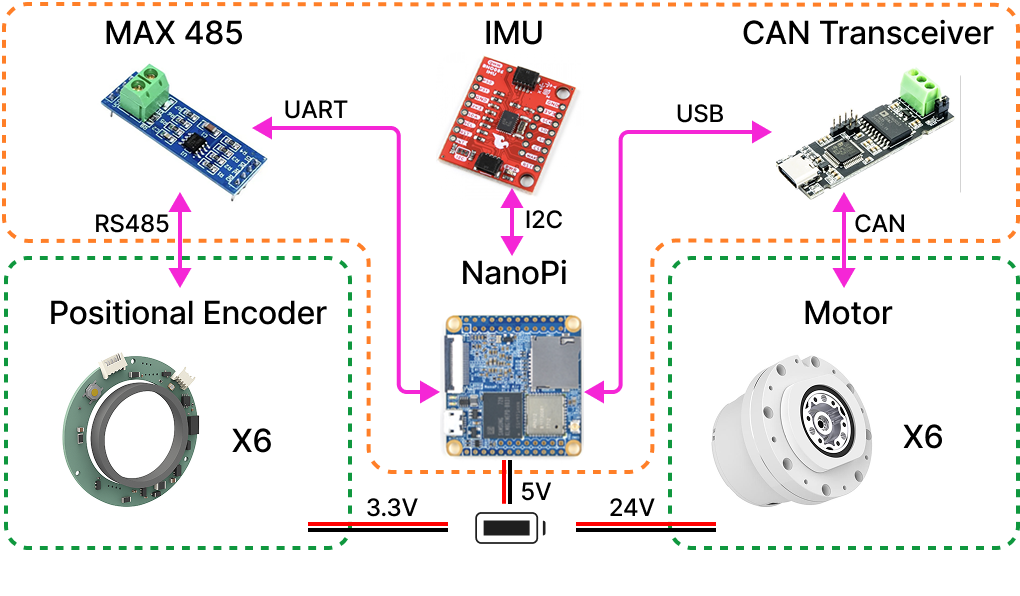}
  \vspace{-22pt}
  \caption{\textbf{The electronics system.}
  The selected mechatronics components are shown.
  The system was designed to be modular and extendable.
  For example, if virtual creatures with hearing or vision are to be realized in future work, microphones and cameras could be easily incorporated into this system with off-the-shelf modules.
  }
  \vspace{-1em}
  \label{fig:electornic_system}
\end{figure}

\textbf{Electronics design.}
To reduce packaging constraints and simplify the electronic-placement problem in the semi-real stage, we adopted a compact onboard computing architecture (Fig.~\ref{fig:electornic_system}). 
The robot uses a NanoPi NEO Air as the main controller, selected for its small footprint to minimize cavity volume requirements. 
Inertial sensing is provided by a BNO086 IMU mounted within the controller enclosure; its pose relative to the body frame is recorded by the pipeline and used for simulation-to-real consistency. 
Each joint is actuated by a Deeprobotics J60-6 motor controlled over a CAN bus via a Makerbase CANable Pro interface. 
Because compliant joint interfaces do not provide reliable mechanical indexing, each joint additionally includes a low-speed positional encoder connected via an RS485 bus for closed-loop feedback.

After printing and hardware integration, all components were assembled according to the automatically generated mounting interfaces (Fig.~\ref{fig:pipeline-overview}K).
No geometry redesign or manual structural modification is required at this stage.
The robot is then switched on (Fig.~\ref{fig:pipeline-overview}L), completing the end-to-end transition from virtual morphology abstraction to a functional physical robot.

\section{Results}
\label{sec:results}

In this section, we measure how well the proposed framework translated a large, diverse set of abstract designs 
into manufacturable blueprints, and how reliably
designs that were successfully translated could be distinguished from those that failed
under the tested hardware constraints. 
We introduce a multipart manufacturability score to quantify the quality of the generated blueprints (Sect.~\ref{subsec:metrics}) and analyze representative failure cases to illustrate which geometric patterns break the pipeline (Sect.~\ref{subsec:success-and-failure}). 
To verify end-to-end functionality, we fabricate a robot based on a blueprint produced by the pipeline,
deploy the robot with a control policy learned in simulation, 
and compare behavior in reality against the simulation (Sect.~\ref{subsec:RL-policy-eval}).
Finally, we test the versatility of the pipeline beyond the tested latent space by replacing the designs decoded from the latent space with a hand-drawn ``sketch'' in voxel space (Sect.~\ref{subsec:sketch2robot}).

\subsection{Evaluation metrics}
\label{subsec:metrics}

To quickly assess how robustly the pipeline handled morphologies uniformly random sampled from the latent space, we measure the pass-through rate, defined as the fraction of sampled robots that successfully complete all solver stages. 
Let $N_{\text{tot}}$ be the number of sampled morphologies and $N_{\text{succ}}$ the number that successfully complete all solver stages. 
Let $N_{\text{fail}}^{\text{stage}}$ denote the number of morphologies whose failure occurs at each stage. 
The pass-through rate is then
\begin{equation}
R_{\text{pass}}
=
\frac{N_{\text{succ}}}{N_{\text{tot}}}
=
\frac{N_{\text{succ}}}{N_{\text{fail}}^{\text{motor}} + N_{\text{fail}}^{\text{elec}} + N_{\text{fail}}^{\text{cable}} + N_{\text{succ}}}.
\label{eq:pass_through_rate}
\end{equation}

To quantitatively evaluate the difficulty of realizing each sampled morphology, 
we defined a series of manufacturability scores. 

A motor feasibility score evaluates whether each joint admits a feasible motor holder–connector configuration under geometric constraints. 
For each joint $j$, we computed a per-joint feasibility score $S_j(\delta_j)$ as defined in Eq.~\ref{eq:motor_score}. 
The optimal offset $\delta_j^\star$ and configuration $c_j^\star$ are obtained by maximizing this volume-based objective.
Then the motor score is defined as the average over all $N_j$ joints:
\begin{equation}
S_{\text{motor}} 
= \frac{1}{N_j} \sum_{j=1}^{N_j} S_j(\delta_j^\star).
\end{equation}
%

An electronics feasibility score measures whether the rigid body provides enough interior clearance to host the electronic module. 
Let $\mathbf{x} \in \mathbb{R}^3$ denote a spatial point and 
$\phi(\mathbf{x})$ the signed distance field (SDF) of the rigid mesh.
We define the maximum interior clearance as $d_{\max} = \max_{\mathbf{x}\,:\,\phi(\mathbf{x})>0} \phi(\mathbf{x})$.
To discourage overly large body parts which are more difficult to 3D print, we further apply a simple size penalty $p_{\text{obb}}\in(0,1]$ computed from the oriented bounding box (OBB) extents. 
The penalty equals $1$ when the extent is below a threshold and decreases only when the body exceeds this threshold. 
The electronics score is then defined as:
\begin{equation}
S_{\text{elec}} = d_{\text{max}}\, p_{\text{obb}}.
\label{eq:elec_score}
\end{equation}

A cable feasibility score assesses the difficultly of cable routing. 
To prevent excessively long cable tunnels and large bending curvature,
which make physical routing difficult,
we compute the total routed length $L_{\text{tot}}=\sum_k L_k$
and the maximum curvature $\kappa_{\max}$ across all cable paths.
The cable score was thus defined as:
\begin{equation}
S_{\text{cable}}
= \exp(-\lambda_L L_{\text{tot}})
+ \alpha\, \exp(-\lambda_\kappa \kappa_{\text{max}}),
\label{eq:cable_score}
\end{equation}
where $\alpha$ balances the penalties on length and curvature,
and $\lambda_L,\lambda_\kappa>0$ are normalization hyperparameters.

Even if individual components are geometrically feasible,
their integration may introduce spatial conflicts that prevent
physical assembly by a human operator. 
In particular, 
(i) electronic modules may
lack sufficient free space for vertical insertion by hand, and
(ii) rigid bodies may interpenetrate after motor integration.
To quantify such assembly-level conflicts,
geometric overlap was penalized using an exponential decay in intersection volume.
Let $V$ denote the corresponding overlap volume.
The electronics installability score was defined as:
\begin{equation}
S_{\text{elec-inst}} = \exp(-V_{\cap}/\lambda),
\label{eq:generic_inst_score}
\end{equation}
where $\lambda>0$ is a normalization constant and $V_{\cap}$ is the 
interference above the module, which indicates whether there is sufficient free space for vertical insertion.
And the
body installability score was defined as:
\begin{equation}
S_{\text{body-inst}} = \exp(-V_{\text{overlap}}/\lambda),
\label{eq:generic_inst_score}
\end{equation}
where $V_{\text{overlap}}$ is the total intersection
volume across all rigid-body pairs.

Together, \(\{S_{\text{motor}}, S_{\text{elec}}, S_{\text{cable}}, S_{\text{elec-inst}}, S_{\text{body-inst}}\}\) provide a stage-wise quantification of manufacturability. We then normalize each term to the same scale, denoted by \(S \in [0,1]\), and define the overall manufacturability score as
\begin{equation}
S_{\text{mfg}}
=
 S_{\text{motor}}
+
 S_{\text{elec}}
+
 S_{\text{cable}}
+
 S_{\text{elec-inst}}
+
 S_{\text{body-inst}}.
\label{eq:overall_mfg_score}
\end{equation}
A morphology is considered fully manufacturable if and only if all solver stages succeed, while the continuous scores enable finer-grained analysis across the latent design space. We do not impose a hard threshold on installability scores.


\begin{table}[!t]
\centering
\caption{Remaining ratio after each stage
for
randomly sampled morphologies (N=1000).}
\begin{tabular}{l c}
\toprule
\textbf{Solver stage} & \textbf{Remaining ratio} \\
\midrule
Motor solver 
& 89.24\% \\
Electronics solver
& 77.85\% \\
Wire solver
& 99.87\% \\
\midrule
Complete pass-through success
& 66.96\% \\
\bottomrule
\end{tabular}
\label{tab:pipeline_stats}
\vspace{-1em}
\end{table} 

\begin{figure*}[!t]
  \centering
  \includegraphics[width=\textwidth]{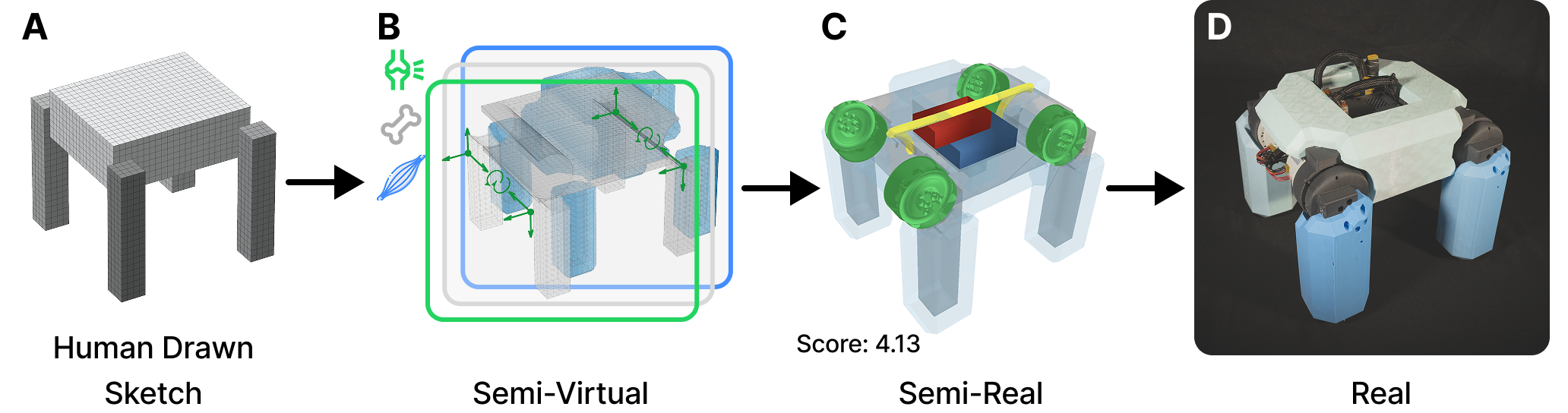}
  \caption{\textbf{Sketch2robot.} 
  Abstract design representations are also convenient for human designers,
  and the pipeline readily accepts a hand-drawn ``sketch'' in voxel space as input (\textbf{A}).
  Voxel workspaces in particular can provide an intuitive and enjoyable interface, as evidenced by the popularity of games like Minecraft.
  Sketches input to the pipeline are automatically processed (\textbf{B}) and embedded with hardware to produce a CAD file (\textbf{C}) that can be directly printed and assembled (\textbf{D}).
  }
  \vspace{-1em}
  \label{fig:quadruped}
\end{figure*}

\subsection{Success and failure modes}
\label{subsec:success-and-failure}

Using the metrics defined above, we evaluated our pipeline on   
$N_{\text{tot}}=1000$ 
test morphologies, which were 
randomly sampled from the latent design space.
The results are shown in Table.~\ref{tab:pipeline_stats}. 
Overall, the pipeline achieved a pass-through rate of $R_{\text{pass}}=66.96\%$, with most failures occurring at the electronics stage, 
fewer at the motor stage, 
and rarely at the wire stage.

Fig.~\ref{fig:multiple_morphologies} shows the test morphologies alongside their corresponding blueprints and the manufacturability score $S_{\text{mfg}}$ defined in Eq.~\ref{eq:overall_mfg_score}. 
High scoring designs (Fig.~\ref{fig:multiple_morphologies}A-F) typically admit feasible motor mounts, 
sufficient internal volume for electronics, 
and low-curvature wire routes, 
providing easy-to-assemble blueprints. 
Low-scoring designs (Fig.~\ref{fig:multiple_morphologies}K-R) exhibited geometric bottlenecks that directly led to solver failures, 
including:
insufficient joint clearance for motor integration (Fig.~\ref{fig:multiple_morphologies}K,L), 
limited internal space for the electronics module (Fig.~\ref{fig:multiple_morphologies}N), 
or lacking a valid kinematic tree for meaningful simulation and policy training (Fig.~\ref{fig:multiple_morphologies}Q,R). 
Overall the scores appear to provide a very good estimate of manufacturability.

To further analyze why certain morphologies fail to pass through the pipeline, 
linear discriminant analysis (LDA) was used to separate the three dominant failure modes along a 2D projection of latent space (Fig.~\ref{fig:failure}A).
The axes of this projection are the first two linear discriminates (LD1 and LD2), which are simply linear combinations of latent features.
The failures were neatly concentrated into three very distinct clusters, which suggests that certain potentially avoidable regions of latent space are more likely to decode into physically infeasible robots than others. 
One cluster (Fig.~\ref{fig:failure}B) corresponded to motor-stage failures, where adjacent rigid segments around a joint were too far apart after semi-virtual processing, leaving insufficient local contact to mount the motor holder/connector. 
Another cluster (Fig.~\ref{fig:failure}C) corresponded to electronics-stage failures, where the rigid bodies provided insufficient internal volume to accommodate the controller and battery modules.
The third and final cluster (Fig.~\ref{fig:failure}D) corresponded to designs that did not possess a valid kinematic tree for simulation. 
Designs that successfully passed through the pipeline without failure 
exhibited variability 
in their overall manufacturability score
(Fig.~\ref{fig:failure}E),
a necessary prerequisite for delineating the most transferrable designs from the least, 
which seems to be the case upon visual inspection
(Fig.~\ref{fig:multiple_morphologies}).

\subsection{Functional testing}
\label{subsec:RL-policy-eval}

To validate end-to-end functionality of our virtual-to-real pipeline, 
we passed an evolved design (Fig.~\ref{fig:pipeline-overview}A)
through the pipeline and followed the resulting blueprint directly without modification to realize the design as a physical robot (Fig.~\ref{fig:pipeline-overview}K).
The policy that was trained for this design in simulation uses a simple two-layer perceptron as the actor, 
which maps a kinematic observation 
to continuous joint position commands. 
The observation includes IMU gravity, IMU angular velocity, motor positions, motor velocities, and the previous action.
The policy successfully transferred to the physical robot in a zero-shot manner: 
the gait and forward velocity exhibited by the physical robot (Fig.~\ref{fig:pipeline-overview}L)
qualitatively matched that of its highly abstract counterpart in simulation.

\subsection{Sketch to robot}
\label{subsec:sketch2robot}

To test the generality of our approach beyond the employed latent space, 
we explored a handcrafted input that, as far as we can tell, does not exist in the latent space (encoding and decoding does not faithfully reconstruct it). 
Although the primary motivation for this work was to facilitate the evolution of robots, we later realized that an automatic pipeline for translating abstract designs into real robots could also have important implications for human-robot interaction:
A user can directly ``paint'' a design by arranging voxels into a desired body shape (Fig.~\ref{fig:quadruped}A). 
The semi-virtual stage processes the rough sketch, converting it into structured rigid/soft meshes and semantic data layers (Fig.~\ref{fig:quadruped}B). 
The semi-real stage embeds motors, electronics, and wire routes to produce fabrication-ready blueprints (Fig.~\ref{fig:quadruped}C). 
We followed the blueprints directly, fabricating and assembling the resulting parts without any CAD redesign, 
realizing the sketch as a physical robot (Fig.~\ref{fig:quadruped}D).


\section{Discussion}
\label{sec:conclusion}

In this paper we introduced a framework for automatically translating freeform virtual creatures into physical robots. 
Unlike other automated procedures for generating manufacturable blueprints,
our approach supports the incremental and open-ended 
evolution of novel robots
because it
does not presuppose
the structure of
the robot's kinematic tree \cite{ringel2025text2robot,xu2021end}
or the shape of its macro-structures~\cite{lipson2000automatic,schaff2022soft, moreno2021emerge, brodbeck2015morphological, yu2025reconfigurable}.
Any body plan that can be ``painted'' into a voxel space 
by a human
(Fig.~\ref{fig:quadruped}A)
or an
evolutionary process (Fig.~\ref{fig:pipeline-overview}A)
can be fed as input to the pipeline.
Certain designs proved to be more manufacturable than others (Fig.~\ref{fig:multiple_morphologies}),
and some designs, of course, could not be realized (Fig.~\ref{fig:failure}) 
under the given set of hardware constraints.
However, we found these failure cases were relatively simple to isolate in the employed latent embedding (Fig.~\ref{fig:failure}), suggesting that they could be
avoided during evolution in simulation 
or penalized when pretraining the latent embedding.

There are several other aspects of the pipeline that could be improved by future work. 
For example, 
the motor solver positioned motors using an intersection-volume heuristic, which only provided an estimate of mechanical strength and did not consider stress concentrations or failure under load.
A better method would be to incorporate structural analysis and optimize mounts for strength and stiffness under task-relevant forces. 
The electronic solver placed electronics at each segment's CoM for simplicity; 
however, in non-convex bodies the CoM can lie outside the volume. 
A better method would be to search over candidate placements within the full rigid volume and explicit optimization for clearance and accessibility. 
The wire solver used surface geodesics as routing references, which restricted channels to near-surface paths and may have missed feasible low-curvature routes through interior free space. 
A better method would be to route wires in 3D while explicitly preserving structural integrity and manufacturability.

However the most important limitation was that the pipeline was specialized to a fixed set of 
components (motor type, controller, battery, wire diameter, etc.) 
and made decisions based on a predetermined set of rules, 
which limited its applicability 
to other design problems with different requirements 
(e.g.~smaller motors for end-effectors). 
Future work will seek to
generalize the virtual-to-real process 
such that hardware is selected dynamically
based on the 
robot's size, shape, material properties, environment, and tasks.

\section*{Acknowledgments}

This research was supported by
NSF awards 2331581 and 2440412,
and
Schmidt Sciences AI2050 grant G-22-64506.

\bibliographystyle{plainnat}
\bibliography{main}

\end{document}